\documentclass[sigconf]{acmart}
\copyrightyear{2021}
\acmYear{2021}
\setcopyright{none}
\acmConference[CIKM '21]{Proceedings of the 30th ACM International
Conference on Information and Knowledge Management}{November 1--5,
2021}{Virtual Event, QLD, Australia}
\acmBooktitle{Proceedings of the 30th ACM International Conference on
Information and Knowledge Management (CIKM '21), November 1--5,
2021, Virtual Event, QLD, Australia}
\acmDOI{10.1145/3459637.3482230}
\acmISBN{978-1-4503-8446-9/21/11}

\usepackage{soul}
\usepackage{hyperref}
\usepackage{inputenc}
\usepackage{graphicx}
\usepackage{amsmath}
\usepackage{amsthm}
\usepackage{amsfonts}   
\usepackage{graphicx}
\usepackage{multirow}
\usepackage{balance}
\usepackage[T1]{fontenc}
\usepackage{listings}
\usepackage{multirow}
\usepackage{color}
\definecolor{codegreen}{rgb}{0,0.6,0}
\definecolor{codegray}{rgb}{0.5,0.5,0.5}
\definecolor{codepurple}{rgb}{0.58,0,0.82}
\definecolor{backcolour}{rgb}{0.95,0.95,0.92}
\definecolor{textblue}{rgb}{.2,.2,.7}
\definecolor{textred}{rgb}{0.54,0,0}
\definecolor{textgreen}{rgb}{0,0.43,0}
\definecolor{codered}{rgb}{201,72,12}

\usepackage[
type={CC},
modifier={by},
version={4.0},
imagewidth=9em
]{doclicense}

\AtBeginDocument{%
  \providecommand\BibTeX{{%
    \normalfont B\kern-0.5em{\scshape i\kern-0.25em b}\kern-0.8em\TeX}}}

\begin{document}
\fancyhead{}

\title{An Efficient Quantitative Approach for Optimizing Convolutional Neural Networks}

\author[Y. Wang et al.]{Yuke Wang$^\dagger$, Boyuan Feng$^\dagger$, Xueqiao Peng$^*$, Yufei Ding$^\dagger$}
\affiliation{
  \institution{\texorpdfstring{$^\dagger$}{dagger}\{yuke\_wang, boyuan, yufeiding\}@cs.ucsb.edu, *\{peng.969\}@osu.edu}
  \institution{\texorpdfstring{{$\dagger$} University of California, Santa Barbara}} \\
  \institution{\texorpdfstring{{*}The Ohio State University}}
  \country{}
 }
  
\begin{abstract}
    With the increasing popularity of deep learning, Convolutional Neural Networks (CNNs) have been widely applied in various domains, such as image classification and object detection, and achieve stunning success in terms of their high accuracy over the traditional statistical methods.
    To exploit potentials of CNN models, a huge amount of research and industry efforts have been devoted to optimizing CNNs.  
    Among these endeavors, CNN architecture design has attracted tremendous attention because of its great potential of improving model accuracy or reducing model complexity.
    However, existing work either introduces repeated training overhead in the search process or lacks an interpretable metric to guide the design. 
    
    To clear these hurdles, we propose \textit{3D-Receptive Field (3DRF)}, an explainable and easy-to-compute metric, to estimate the quality of a CNN architecture and guide the search process of designs. 
    To validate the effectiveness of \textit{3DRF}, we build a static optimizer to improve the CNN architectures at both the stage level and the kernel level. 
    Our optimizer not only provides a clear and reproducible procedure but also mitigates unnecessary training efforts in the architecture search process.
    Extensive experiments and studies show that the models generated by our optimizer achieve up to 5.47\% accuracy improvement 
    and up to 65.38\% parameters deduction, compared with state-of-the-art CNN model structures like MobileNet and ResNet.
\end{abstract}



\maketitle
\doclicenseThis

\section{Introduction} \label{sec:intro}
Deep convolutional neural networks (CNNs) have achieved significant successes in a broad collection of fields, including object-detection \cite{object_detection}, video classification \cite{Karpathy_2014_CVPR}, object tracking~\cite{object_tracking}, image segmentation~\cite{segmentation} and human pose estimation~\cite{human_pose}. Such unparalleled successes attract many interests in CNN architecture design to \textit{improve accuracy} or \textit{reduce complexity}.
Examples include an array of efficient models that have been crafted manually (\textit{e.g.,} VGG \cite{vgg}, MobileNet \cite{mobilenet_2017_howard}, ShuffleNet \cite{ma2018shufflenet}) and those generated automatically by the neural architecture search (NAS) tools~\cite{real2017large,Liu_2017_ICCV,liu2018progressive,baker2016designing,zoph2018learning}.
Yet, two challenges of CNN architecture design remain far from well resolved: 1) missing an interpretable metric, and 2) huge training efforts. The former indicates that some direct and easy-to-interpret metric is still missing to guide the design, while the latter means that the repeated training cost is huge for evaluating different architectures in the search process.

To address these challenges, we propose \textit{3D-Receptive Field (3DRF)}, an interpretable metric, for efficient CNN architecture designs. 
Particularly, we focus on two levels: the \textit{stage level}\footnote{Following many works~\cite{he2016deep,shufflenet_zhang_2017,zhang2018clcnet}, we define a stage in a CNN as a collection of consecutive convolution layers with \textit{input tensors} of the same spatial dimensions (\textit{i.e.,} pooling or convolution kernel with stride $\geq$ 2 will generate a new stage).} and the \textit{kernel level}. 
At the stage level, we decide the number of convolution kernels in different stages, while at the kernel level we choose the type of the convolution kernel to use (\textit{i.e.,} standard convolution kernels or efficient factorized kernels \cite{depthwise_origin,shufflenet_zhang_2017}).
We build up \textit{3DRF} to uniformly conduct the optimization at both levels. 
The key insight is that the portion of the input tensor that can flow into each output neuron, which we name as \textit{3DRF}, often determines the learning potential of that given stage or kernel. 
A stage or kernel with larger \textit{3DRF} will have more input elements passing through, leading to a higher potential for extracting useful features and improving the classification accuracy.
Therefore, we use \textit{3DRF} to estimate the quality of architecture design in the search process, rather than repeated training.

To validate and showcase the effectiveness of \textit{3DRF}, we propose an \textit{architecture optimizer} to examine CNN architecture designs at stage and kernel level. 
At stage level, we provide an \textit{organizer} to improve the accuracy of a CNN model while using the same or fewer convolution kernels. The organizer, in effect, removes the convolution kernels that cannot contribute to \textit{3DRF} enough or move the kernels from the positions with marginal contributions to \textit{3DRF} in one stage to another stage with larger contributions. The optimization is based on two key observations: 1) the contributions from the latter kernels in a stage are diminishing since the newly observed input elements are on the marginal positions, which have less impact compared with the central input already observed; 
2) when the spatial size of the input tensor to a stage is small, piling more layers can barely learn more features. On the other side, moving some layers to another stage with larger input tensor would promote \textit{3DRF} and better learning capacity.

At the kernel level, we propose a \textit{decomposer} to reduce model complexity without substantively affecting accuracy.
The decomposer, in effect, replaces standard convolution kernels~\footnote{In this paper, we refer to the standard convolution kernel as the one with 3 * 3 * C filters, where C is the number of input channels.} with convolution blocks composed of efficient factorized kernels (\textit{e.g.,} Depthwise Convolution~\cite{depthwise_origin}, and Pointwise Convolution~\cite{inceptionV4}). 
The key guidance behind such replacement is to maintain the same \textit{3DRF} (\textit{i.e.,} the efficient convolution block should observe the same amount of \textit{3DRF} as standard convolutions in order to maintain accuracy). We name this rule as \textit{Rule for Kernel Replacement}. 
This rule not only allows us to unify all existing convolution blocks used in MobileNet, ShuffleNet, clcNet~\cite{zhang2018clcnet}, and Xception~\cite{xception}, but also inspires the discovery of one new basic factorized convolution kernel, as we named \textit{Rolling Pointwise Convolution (RPW)}, 
and a new convolution block (Depthwise (DW) + RPW).  
This new \textit{convolution block} turns out to be more efficient than existing factorized kernel designs, like that in MobileNet model. 

To facilitate the end-to-end CNN model design, we introduce our design prototype. As shown in the Listing~\ref{example code for 3DRF optimizer}, we start with importing our 3DRF-based optimization libraries, including a stage optimizer (\texttt{stage\_opt}) and a kernel optimizer (\texttt{kernel\_opt}). We will then build a CNN models as we normally do in the regular Pytorch. Here, convolutional layers in the CNN models can be grouped into different stages, where each stages consists of convolutions linearly stacked together. Different stages are sequentially connected. At the end of those stages, we put the linear (fully-connected) layer and a softmax layer layer to generate logits for classification.	\setlength{\textfloatsep}{5pt}
\begin{figure}[t] \small
\centering
\begin{lstlisting}[caption=Illustration of 3DRF-based Optimizer Prototype., label=example code for 3DRF optimizer]
from 3DRF_optimizer import stage_opt, kernel_opt
# import other libraries, such as Pytorch...

# Create an stage of CNN model.
def make_stage(stage_depth):
    layers = nn.sequential()
    for i in range(stage_depth):
        layers.append(nn.conv2D(inChannel, outChannel))
    return layers
    
# Create a CNN model.
class CNN(nn.module):
   def __init__(self, stageDepth=[2,2,2,2], outClass=10):
        self.stages = torch.nn.moduleList()
        for depth in stageDepth:
            self.stages.append(make_stage(depth))
        self.classifier = nn.Linear(flatDim, outClass)
        self.softmax = nn.softmax()
        
    def forward(self, X):
        out = X
        for stg in self.stages:
            out = stg(out)
        out = self.classifier(out)
        out = self.softmax(out)
        return out
# Define a simple CNN Model.
model = CNN([2,2,2,2], 10)
# Compute the delta 3DRF for a input model.
info_3DRF = stage_opt.comp_Delta3DRF(model)
# Optimize the model structure with delta 3DRF.
model_opt = stage_opt.optimze_arch(model, info_3DRF)
# Optimize the kernel.
model_final = kernel_opt(model_opt)
# Do regular model training and inference.
\end{lstlisting} 
\label{code: example code for 3DRF optimizer.}
\end{figure}

In summary, the major contributions of our work are:
\begin{itemize}
    \item We propose a brand-new interpretable metric \textit{3D-Receptive Field (3DRF)} for guiding CNN architecture designs efficiently. Whereas previous CNN model architecture exploration techniques (\textit{e.g.}, NAS) require huge training and searching efforts. 
    \item We build an end-to-end CNN stage-level organizer for improving the accuracy performance of CNN models at the model architectural level. This can largely ease the manual efforts in arduous CNN model optimization process.
    \item We introduce an new type of convolution kernel -- Rolling-Pointwise Convolution to reduce the model parameters and the computation FLOPs.
\end{itemize}

Rigorous evaluations on real-world image datasets (\textit{e.g.}, CIFAR-10/100~\cite{cifar10}, and ImageNet~\cite{deng2009imagenet}), demonstrate the strength of our architecture optimizer in terms of model accuracy, FLOPs and parameters. 
At the stage level, the organizer improves the accuracy (up to 5.47\%) of the manually crafted CNN structures (\textit{e.g.}, MobileNet) by maximizing the contribution to \textit{3DRF}. 
For instance, the optimized MobileNet achieves 3.7\% higher accuracy with 74\% fewer parameters and 16\% fewer FLOPs compared with the original structure. 
At the kernel level, the newly discovered convolution block achieves higher accuracy (up to 0.58\%) with much fewer computations (up to 40.0\% reduction) and parameters (up to 90.4\% reduction) compared with the existing design. 
For example, one kernel designed by us has 2.54\% higher accuracy and 29.04\% fewer FLOPs in comparison with the MobileNet. 
\begin{figure*} [t] \small
    \centering
    \includegraphics[width=\textwidth]{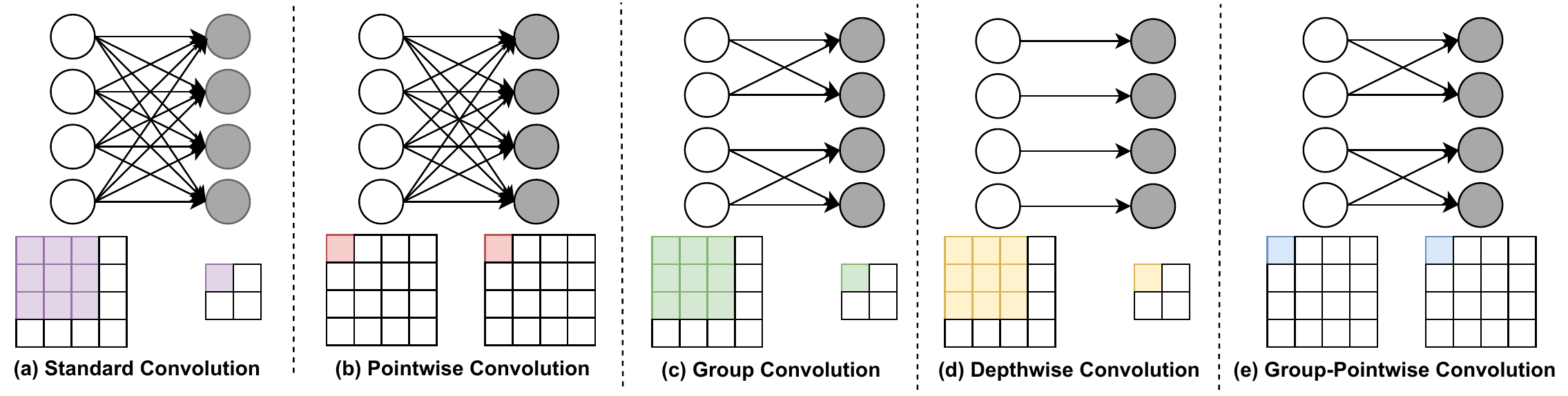}
     \vspace{-10pt}
    \caption{Channel mapping (top) and Spatial mapping (bottom) of the standard convolution and factorized convolution kernel.}
    \label{fig: Comparison of different convolutional kernels.}
\end{figure*}

\section{Related Work} 
\label{sec:related_work}

\subsection{Neural Architecture Search (NAS)}
NAS methods have been widely studied to automatically construct efficient CNN architectures. 
NAS frameworks generally come with three major components, 
1) \textit{Search space:} 
The NAS search space is composed of several types of operations (\textit{e.g.}, convolution, fully-connected, and pooling) and the inter-connection among these operators. 
The design of search space demands domain expertise from both the deep learning and the specific application settings; 
2) \textit{Search algorithm:} 
A NAS search algorithm samples a population of network architecture candidates. It receives the model performance evaluation result (\textit{e.g.,} result) as rewards
and optimizes to generate high-performance architecture candidates.
3) \textit{Evaluation strategy:} 
This step will measure the performance of candidate models in order to improve the search algorithm. 

The most significant part of NAS research has been devoted to the neural architecture search algorithm. 
And a array of techniques and strategies have been proposed, such as evolutional algorithms~\cite{real2017large,real2018regularized}, hill climbing~\cite{elsken2017simple}; multi-objective search~\cite{elsken2018efficient,zhou2018neural}, and reinforcement learning (RL)~\cite{real2017large,Liu_2017_ICCV}. 
To accelerate the NAS search, ENAS~\cite{pham2018efficient} represents the search space using a directed acyclic
graph (DAG) and targeting at optimizing the subgraph structure within the large supergraph. 
Meanwhile, it also introduces a training strategy of parameter sharing among subgraphs to significantly boost the searching efficiency. 
Work from~\cite{jin2019rc, liu2018darts} also follow the similar idea of hierarchical computation graph optimization.
Work from~\cite{yu2019scaling} further share the parameters of different paths within a block using super-kernel representation. 
\cite{Mei2020AtomNAS} proposes a fine-grained search space comprised of atomic blocks that is much smaller than the ones used in recent NAS algorithms. 

Although NAS methods can build high-quality CNN architecture, they have two major drawbacks. 
First, they require prohibitively expensive computing power and add significant overhead to the design time. For instance, the RL-based method in~\cite{zoph2018learning} requires 500 NVIDIA P100 GPUs for more than 4 days to evaluate 20000 candidate neural networks, even after adopting many proxy tasks techniques including early stopping with few epochs, running on a small dataset, and limiting the kernel numbers. 
Second, the NAS method can identify the design, but it does not explain the general rule behind to obtain such a design, which limits its applicability. Once the task changes, one has to run NAS again. In contrast, our static architecture optimizer gives an alternative solution, offering a clear and reproducible design procedure without training in the architecture search process. Other works still requires non-trivial overhead of CNN runtime profiling for optimization.

\subsection{Standard Convolution}
The widely applied deep learning application demands effective ways to capture the characters of the inputs (\textit{e.g.}, images). Among those techniques, the standard convolution is most widely used in many CNNs~\cite{resnext, vgg, sandler2018mobilenetv2}.
In general, we annotate the input image ($I$), output feature map ($O$), and filter ($F$). The dimension of an image is $[I_w, I_w, C_{in}]$, where $I_W$ is the size of an image while $C_{in}$ is the number of input channels (\textit{e.g.}, the RGB image has 3 input channel). The standard convolution (Figure~\ref{fig: Comparison of different convolutional kernels.}a) leverages $C_{out}$ standard convolutional filters with the shape of $[K, K, C_{in}]$, where the $K$ is the filter size, $C_{in}$ is the number of input channels, and $C_{out}$ is the number filters. After applying the standard convolution on the input (with the shape of $[I_{w}, I_{w}, C_{in}]$), we will get the output feature map $O$, which has the shape of $[O_{w}, O_{w}, C_{out}]$., where the $O_{w}$ is size of the output feature map. 
Note that the mainstream CNNs~\cite{resnext,vgg,mobilenet_2017_howard} generally maintain the same feature map spatial dimension at different convolutional layers while only changing the number of the channels across different layers. 

Formally, for standard convolution, we have
\begin{equation}
    O_{m, n, c} = \sum^{K, K, C_{in}}\limits_{i,j,a}F_{i,j,a,c} * I_{m+i-1,n+j-1,a}
\end{equation}
where $O_{m,n,c}$ is one pixel point in the output feature map; $m$ and $n$ are the spatial indexes in the output feature map ($m\in Z: m\in[0, O_w)$ and $n\in Z: n\in[0, O_w)$); 
$a$ is the channel index in the input feature map ($a\in [0, C_{in})$);
$c$ is the channel index in the output feature map ($c\in Z: c\in [0, C_{out})$); $i$, $j$, and $a$ are the index used to accumulated the elementwise multiplication values between input feature map and one filter.
The standard convolution will not only extract the spatial information by traversing a $K\times K$ 2D sliding window within each channel but also effectively fuses the information across different channels (Figure~\ref{fig: Comparison of different convolutional kernels.}a), where each kernel filter will gather the information from all input channels.

\subsection{Kernel Factorization}
Besides the standard convolution kernel, recent deep-learning research introduces several factorized kernels~\cite{inceptionV4,alexnet_2012,depthwise_origin,shufflenet_zhang_2017} and combine them into a convolution block. This can offer another way to improve the computation efficiency of CNN architecture designs while maintaining the prediction power. 
Existing factorized kernels can be divided into four categories. 
Specifically, the first type is the Pointwise Convolution (PW)~\cite{inceptionV4}) (Figure~\ref{fig: Comparison of different convolutional kernels.}b), which is a standard convolution with $1\times1$ spatial size. 
The second type is Group Convolution~(GC)~\cite{alexnet_2012} (Figure~\ref{fig: Comparison of different convolutional kernels.}c) that divides input channels into several groups and performs standard convolution within each group. 
The third type is Depthwise Convolution (DW)~\cite{depthwise_origin} (Figure~\ref{fig: Comparison of different convolutional kernels.}d) which calculates spatial convolution per channel or can be regarded as an extreme case of GC when the group number equals the number of the input channels. 
The last one is Group Pointwise Convolution~(GPW)~\cite{shufflenet_zhang_2017} (Figure~\ref{fig: Comparison of different convolutional kernels.}e), that further splits PW into groups.
Previously, researchers combine some of the factorized kernels into convolution blocks. 

Xception~\cite{xception} and MobileNet~\cite{mobilenet_2017_howard} demonstrate the successful application of convolutional kernel factorization in the popular CNN models. 
It breaks the original standard convolution into two parts: \textbf{depthwise} (DW) convolution and \textbf{pointwise} (PW) convolution. The first step (DW) applies $C_{in}$ different $[W, W, 1]$ filters to each of the $C_{in}$ input channels independently, which can be formalized as Equation~\ref{equ: depth-wise convolution}
\begin{equation} \label{equ: depth-wise convolution}
\hat{O}_{m,n,a} = \sum^{K,K}\limits_{i,j}F^{(dw)}_{i,j,a}*I_{m+i-1, n+j-1,a}
\end{equation}
The second step (PW) applies a filter with $1\times1$ spatial dimension. 
As shown in Equation~\ref{equ: point-wise convolution}.
\begin{equation} \label{equ: point-wise convolution} 
O_{m,n,c} = \sum^{C_{in}}\limits_{a}F^{(pw)}_{a,c}*I_{m-1,n-1,a}
\end{equation}
In this paper, we use the idea of \textit{3DRF} to unify these previous convolution blocks. In addition, we create a new type of factorized convolution kernel, named \textit{Rolling Pointwise Convolution (RPW)}, and a new convolution block (DW+RPW) that can outperform the previous designs.

\section{3D-Receptive Field} \label{sec:definition}
In this section, we present \textit{3D-Receptive Field} (\textit{3DRF}) for measuring the representation ability of each neuron in a convolution layer.
Then, we derive the \textit{3D-Receptive Field Gain} (\textit{3DRF} Gain) for quantifying the representation ability change when an additional convolution layer is inserted.
This \textit{3DRF} Gain is sensitive to the location, type, and combination of the inserted convolution layer, thus guiding the CNN design.
We demonstrate the effectiveness of \textit{3DRF} Gain in quantifying representation ability, in terms of its impact on accuracy.

Our 3D-Receptive field is inspired by an existing metric,  \textit{receptive field} \cite{sherrington1906observations}, which quantifies the spatial area of neurons for evaluating a single neuron in the next convolution layer.
This receptive field serves well for quantifying the local representation ability in a single traditional convolution layer, where a larger receptive field leads to higher accuracy.
However, the receptive field fails to quantify the global representation ability across layers, when a large number of convolution layers with diverse receptive fields stacked in a CNN stage.
Moreover, the receptive field fails to consider the channel number, which becomes critical in modern convolution layers (\textit{e.g.}, Depthwise convolution and Channel-wise convolution).
By contrast, our \textit{3DRF} provides the first global metric for quantifying the global representation ability across layers, considering extensively the location, type, and combination of convolution layers.
By quantifying the global representation ability, \textit{3DRF} serves as an effective and efficient tool for guiding the CNN design without tediously enumerating and training NN architectures.

\begin{figure}[t]
    \centering
    \includegraphics[width=\columnwidth]{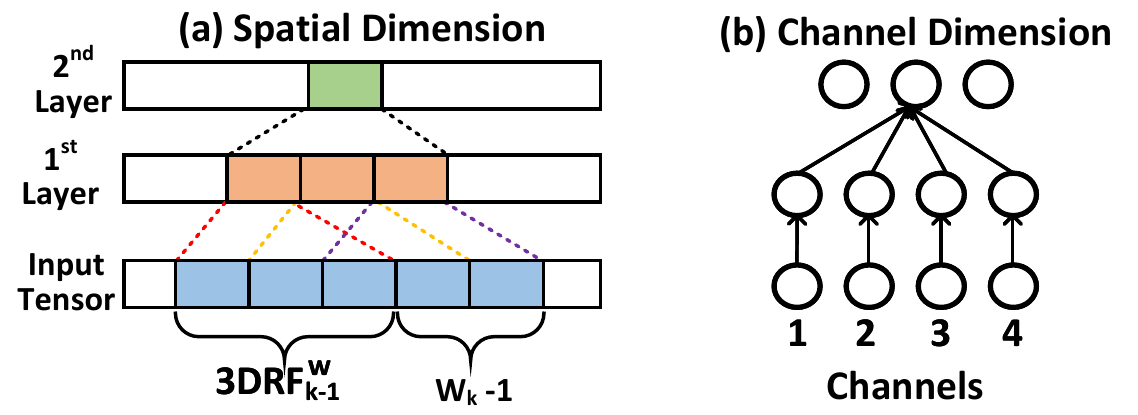}
    \caption{Illustration of 3D-Receptive Field (\textit{3DRF}) for convolutions of a single stage.}
    \label{fig:IFDef}
\end{figure}

\subsection{Definition of 3D-Receptive Field}
For a CNN stage with a sequence of layers, we define the 3D-Receptive Field (\textit{3DRF}) for the $k^{th}$ convolution layer in the current stage as $3DRF_k$.
This $3DRF_k$ captures the number of neurons in the initial input tensor to the CNN stage that contributes to computing individual neurons in this layer $k$.
This initial input tensor is the $w_0\times w_0 \times 3$ input tensor (\textit{e.g.}, input image) in the first stage of a CNN, and a $w_0\times w_0 \times c_0$ input tensor in later stages.
Here, $w_0$ is the spatial width of the input tensor and $c_0$ is the channel number of the input tensor.
To cater convolution layers with diverse kernel sizes and types, \textit{3DRF} considers two factors of the spatial width $3DRF_k^w$ for the kernel size and the channel number $3DRF_k^c$ for the convolution type:
\begin{equation} \label{equ:1}
     3DRF_{k} = (3DRF_k^w)^d * 3DRF_k^c
\end{equation} 
where $d=1$ for 1D convolution (Figure~\ref{fig:IFDef}) and $d=2$ for 2D convolution. We recursively compute the spatial width $3DRF_k^w$ in layer $k$ based on the spatial width $3DRF_{k-1}^w$ in the preceding layer $k-1$ and the kernel width $w_k$ in the current layer $k$:
\begin{equation} \label{eq:IF_k}
    3DRF_k^w = \min(3DRF_{k-1}^w + w_{k} - 1, w_0)
\end{equation}
A $min()$ is applied for ensuring that the spatial width $3DRF_k^w$ does not exceed the spatial width $w_0$ of the input tensor.

We compute recursively the channel number $3DRF_k^c$ in layer $k$ with a property function $g(\cdot, \cdot)$, that captures the channel number $3DRF_{k-1}^c$ in the preceding layer $k-1$ and the convolution type $T_k$ in the current layer $k$:
\begin{equation} \label{equ:IF_c}
    3DRF_k^c = \min(g(3DRF_{k-1}^c, T_k), c_0)
\end{equation}
A $min()$ is applied for ensuring that the channel number $3DRF_k^c$ does not exceed the channel number $c_0$ of the input tensor. 
The property function $g(\cdot, \cdot)$ captures the information flow from the perspective of channel numbers and is designed for individual convolution types.
For example, as illustrated in Figure~\ref{fig:IFDef}, we set the property function $g(3DRF_{k-1}^c, PW) = c_0$ for Pointwise (PW) Convolution, since the output neuron of PW observes all input channels.
Similarly, we set $g(3DRF_{k-1}^c, DW) = 3DRF_{k-1}^c$ for Depth-wise Convolution (DW), since only one channel from the preceding layer $k-1$ contributes to the neuron in the current layer $k$.
This property function $g(\cdot, \cdot)$ is designed only once for a small set of convolution types.
While modern CNNs may have hundreds of convolution layers, these layers often use the same convolution type repeatedly.
Thus, the property function can be written once and applied repeatedly for a large number of convolution layers.

\vspace{-5pt}
\subsection{Definition of \textit{3DRF} Gain}
We derive the \textit{3DRF} Gain ($\Delta 3DRF$) to measure the impact of a convolution layer $k$ over the model representation ability, in terms of the impact over the 3DRF.
While \textit{3DRF} quantifies the information flow in a convolution unit as a whole, \textit{3DRF Gain}---denoted by $\Delta 3DRF$--- targets at measuring the contribution of a single convolution kernel $k$ in the unit. 
The goal of introducing $\Delta 3DRF$ is to create a direct indicator that could match the learning power (\textit{i.e.,} prediction accuracy) of a CNN model in the granularity of a single convolution, laying a foundation for static architecture optimization. 
Specifically, we define $\Delta 3DRF_{k}$ as the difference in the receptive field with and without the layer $k$, adjusted with an exponential decay term:
\begin{equation} \label{equ:3}
     \Delta 3DRF_k = 
     \frac{3DRF_k - 3DRF_{k-1}}{3DRF_{k-1}} * e ^ {-\alpha * \frac{3DRF_{k-1}}{V_0}}
\end{equation}
where $V_0 = w_0 \times w_0 \times c_0$ is the volume of the input tensor. 
The exponential decay term rescales the impact of the $k^{th}$ layer with regards to the information already observed by $1^{th}$ to $(k-1)^{th}$ layers. 
which composes of two major terms: the former calculates the relative increase in \textit{3DRF} incurred by kernel $k$; the latter introduces an exponentially decay term to rescale the impact of the $k^{th}$ layer with regards to the information already observed by $1^{th}$ to $(k-1)^{th}$ layers.
This decay term is inspired by the observation that the elements in the central region of the input tensor usually have a larger impact than the newly observed elements on the margin: the central input elements have more paths to propagate their values into the output in the forward pass and larger gradient in the backward pass.
Note that $\alpha$ is a hyperparameter that should be set larger than 0. In our empirical study, we tried multiple choices and observed no substantial difference in architecture optimization, and we set it to 3 for the rest of this paper. 
\begin{table}[t] \small
    \caption{\small Illustration of computing \textit{3DRF} Gain on Variant-3.}
    \label{tab:vgg_IF}
    \centering
    \scalebox{1}{
    \begin{tabular}{cccccc}
        \toprule
        k & Layer Type  & $3DRF_k^w$ & $3DRF_k^c$ & $3DRF_k$ & $\Delta 3DRF_k$\\
        \midrule
        \midrule
        1 & conv3-256  & 3 & 128 & 1152 & -\\
        2 & conv3-256 & 5 & 128 & 3200 & 1.17\\
        \midrule
        3 & conv3-256 & 7 & 128 & 6272 & 0.29\\
        \bottomrule
    \end{tabular}}
    \vspace{-5pt}
\end{table}
\begin{table}[t]
    \caption{Impact of \textit{3DRF} Gain ($\Delta 3DRF$) over Accuracy.}
    \vspace{-5pt}
    \label{tab:vgg_stages}
    \centering
    \scalebox{1}{
    \begin{tabular}{cccc}
        \toprule
        Network  & $\Delta 3DRF$ & Accuracy (\%) & $\Delta$Accuracy~(\%)\\
        \midrule
        \midrule
        VGG-11  & 0 & 92.68 & 0 \\
        \midrule
        Variant-1 & 1.73 & 93.56 & 0.88\\
        Variant-2 & 1.60 & 93.46 & 0.78\\
        \midrule
        Variant-3 & 0.29 & 92.75 & 0.07\\
        \midrule
        Variant-4 & 0.0 & 92.58 & -0.10\\
        Variant-5 & 0.0 & 92.41 & -0.27\\  
        \bottomrule
    \end{tabular}}
\end{table}
\begin{figure*}[t]
    \centering
    \includegraphics[width=0.77\linewidth]{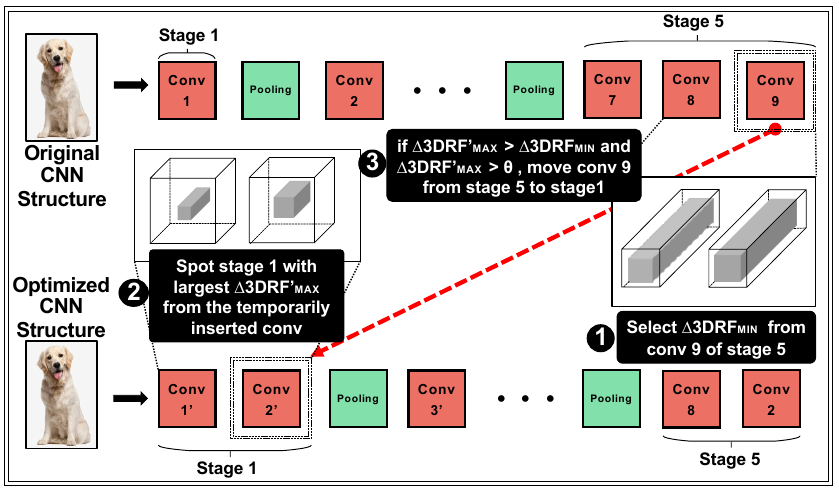}
    \vspace{-5pt}
    \caption{Illustration of the Stage-level Organizer.} 
    \vspace{-9.5pt}
    \label{fig:organizer}
\end{figure*}

\subsection{Case Study: Accuracy Impact of \textit{3DRF} Gain}
We demonstrate the impact of diverse ($\Delta 3DRF$) over the accuracy.
Here we generate diverse ($\Delta 3DRF$) by sticking to the same base model and inserting an additional convolution layer at diverse location.
More study on the ($\Delta 3DRF$) from varying the type and combination of convolution layers will be conducted later in the evaluation section.
As shown in Figure~\ref{tab:vgg_stages}, we take VGG11~\cite{vgg} as the baseline structure and run it on CIFAR-10 dataset~\cite{cifar10}. Specifically we generate five VGG-variants by inserting a single standard convolution before each max pooling.
The inserted convolution layer has the same kernel width and channel number as its preceding layer.
For example, we insert a conv3-64 before the first max pooling as the Variant-1, and a conv3-512 before the fifth max pooling as the Variant-5. 
Specifically, we train these models on the CIFAR-10 training dataset and report the accuracy on the CIFAR-10 testing dataset.
We repeat this procedure for ten times and present the average accuracy here.
We also present the $\Delta 3DRF$ of each variants for demonstrating the impact of $\Delta 3DRF$ over accuracy.
$\Delta 3DRF$ is calculated by leveraging our proposed Equation~\ref{equ:3} for the newly inserted layer.

As shown in Table~\ref{tab:vgg_IF}, the procedure of computing $\Delta 3DRF$ on Variant-3, which inserts an additional layer to the third stage in VGG-11.
Originally, the third stage in VGG-11 contains two convolution layers (\textit{i.e.}, the $1^{st}$ layer and the $2^{nd}$ layer in Table \ref{tab:vgg_IF}).
We insert the $3^{rd}$ convolution layer with the same kernel width and channel number as the first two layers.
The input tensor to this third stage is of shape $8\times 8 \times 128$, leading to a $V_0$ of $8192$.
Following Equation \ref{equ:1} - \ref{equ:IF_c}, we can compute $3DRF_k^w$, $3DRF_k^c$, and $3DRF_k$ recursively.
The derived $3DRF_k$ can be exploited for computing $\Delta 3DRF$ following Equation \ref{equ:3}.
This procedure can be applied for other VGG-11 model variants, leading to the $\Delta 3DRF$ in Table \ref{tab:vgg_stages}.

As shown in Table~\ref{tab:vgg_stages}, we can clearly figure out the the impact of $\Delta 3DRF$ on CNN model accuracy.
Large $\Delta 3DRF$ of the newly inserted layer agrees with notable accuracy gain, as is the case for \textit{Variant-1} and \textit{Variant-2}.
For the \textit{Variant-3}, small $\Delta 3DRF$ indicates close-to-saturation information coverage, yielding negligible accuracy improvement from the original model.
Variant-4 and Variant-5 has a low $\Delta 3DRF$ of $0$, indicates that inserting convolution layers does not improve its \textit{3DRF}.
The insight is that, for an input tensor with a small spatial width $w_0$ of $2$ (after 4 times of max pooling from an input image of shape $32 \times 32 \times 3$), a single convolution layer of kernel width $3$ is sufficient for capturing all neurons.
In fact, Variant-4 and Variant-5 show an accuracy degradation of $-0.10\%$ and $-0.27\%$ respectively.
This degradation shows that a $\Delta 3DRF$ of $0$ signals overfitting since all input elements have already been observed by other kernels at such stage.
Comparing across variants, Variant-1 has a larger $\Delta 3DRF$ of $1.73$ and a larger $\Delta Accuracy$ of $0.88\%$, compared with Variant-5 with $\Delta 3DRF$ of $0.0$ and $\Delta Accuracy$ of $-0.27\%$.
This trend demonstrates a strong correlation between the $\Delta 3DRF$ and the $\Delta Accuracy$, thus guiding the NN design in terms of the insertion location.

To sum up, $\Delta 3DRF$ effectively probes the potential of accuracy improvement, and we leverage such an easy-to-compute metric to build our architecture optimizer in Section~\ref{sec:design}.
\section{Architecture Optimizer via 3DRF} \label{sec:design}
We build a static \textit{Architecture Optimizer} based on \textit{3DRF} and $\Delta 3DRF$. 
It examines the structure inefficiency in a given CNN architecture and optimizes it at the \textit{stage level} and \textit{kernel level}. 

\subsection{Stage-Level Organizer}
\label{subsec:organizer}
Stage-level organizer (Figure~\ref{fig:organizer}) manages to improve the prediction accuracy of a CNN design by iteratively removing a convolution kernel from a saturated stage or moving it to another stage with more room to absorb new information (\textit{i.e.,} learn from more marginal elements introduced by the kernel).

Three sub-steps are conducted in each iteration.
The first step is to find the convolution kernel with minimum $\Delta 3DRF$, which has the lowest contribution to the 3DRF. 
In consideration of the decaying property of $\Delta 3DRF$ within a stage, this step can be simplified to compute the $\Delta 3DRF$ of the last convolution kernel in each stage.
Comparing across stages, we select the convolution layer with the minimum \textit{3DRF} Gain, denoted as \textit{ $\Delta 3DRF_{MIN}$} in Figure~\ref{fig:organizer}, and identify the corresponding stage as the \textit{source stage}.
This identified convolution layer will be either deleted or moved from the source stage to another stage, in the following steps.

The second step is to spot the stage with the largest room for improving 3DRF.
This step follows the insight from our case study that a larger $\Delta 3DRF$ often leads to higher accuracy.
We tentatively append the convolution kernel identified in the first step to each stage and compute the corresponding $\Delta 3DRF$.
When appending the convolution layer, the input and output channel number will be adjusted for catering to the preceding layers in the source stage and the following layer in the next stage if available.
Comparing across stages, we can find the one, called \textit{target stage}, with maximum $\Delta 3DRF$ for the appended layer (\textit{$\Delta 3DRF'_{MAX}$} in Figure~\ref{fig:organizer}). 
This step follows the insights obtained from our case study that a strong correlation exists between $\Delta 3DRF$ and $\Delta Accuracy$.
to conduct architecture optimization. 

The third step decides whether moving the last convolution layer from the source stage to the target stage or simply removing this layer.
When moving the convolution layer, we adjust the input channel number and the output channel number with the same strategy in the second step.
This step follows the insights obtained from our case study to conduct architecture optimization. 
There are three key choices: 1) If \textit{$\Delta 3DRF'_{MAX}$ > $\Delta 3DRF_{MIN}$ and $\Delta 3DRF'_{MAX}$ > $\theta$}, we move the last kernel from the \textit{source stage} and append it to the \textit{target stage};  2) If \textit{$\Delta 3DRF'_{MAX}$ < $\theta$ and $\Delta 3DRF_{MIN}$ < $\theta$}, we just remove the last kernel from the \textit{source stage} (no appending); 
3) If $\Delta 3DRF_{MIN}$ > $\Delta 3DRF'_{MAX}$ and $\Delta 3DRF_{MIN}$ > $\theta$, we keep the original structure and terminate our optimization procedure.  
Here the hyperparameter $\theta$ is the border we draw empirically to distinguish underfitting from overfitting. 
For example, $\theta$ is set to 0 for VGG. 
Following this iterative optimization procedure, our organizer manages to mitigate the structure-level inefficiency in a CNN design via static architecture optimization. The experimental results of the organizer can be found in our evaluation.
\begin{figure*}[ht!]
    \centering
    \includegraphics[width=\linewidth]{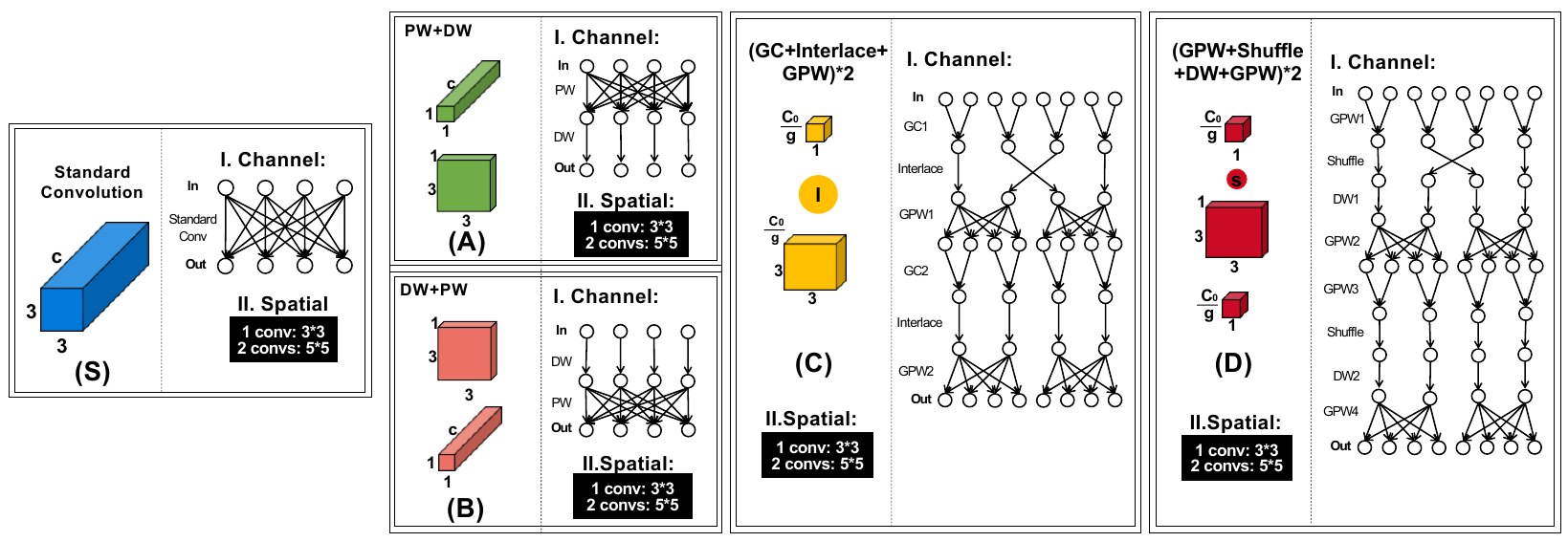}
    \vspace{-20pt}
    \caption{Illustration of the \textit{3DRF}, both in the channel (I) and  spatial (II) dimension, for the standard kernels ($S$) and previous convolution blocks~($A$-$D$). $g$ is the number of groups for GC and GPW. The arrow denotes the flow from inputs to outputs in the channel dimension, and the number of input channels that could flow into an output neuron would be the channel dimension of \textit{3DRF} for that block. We omit the process of computing the spatial size of \textit{3DRF}, while only giving the computed result based on Equation~\ref{equ:1} in the figure.} 
    \label{fig:my_label}
\end{figure*}

\subsection{Kernel-Level Decomposer}
\label{subsec:factorization}
At the kernel level, our \textit{decomposer} reduces the computational cost of a CNN architecture design, by substituting its \textit{standard convolution kernels} with less computational expensive \textit{convolution blocks}. The key challenge here is to construct such an efficient and effective \textit{convolution block} with multiple factorized kernels. Previous manual efforts by domain experts have made some progress~\cite{sandler2018mobilenetv2,zhang2018clcnet,shufflenet_zhang_2017}, but the underlying design principle remains unclear. 
In this paper, we provide the first easy-to-follow design principle, \textit{Rule of Kernel Replacement}, to guide the design of efficient convolution blocks.

\vspace{2pt}
\textbf{Rule of Kernel Replacement}
To avoid significant accuracy degradation and achieve computation efficiency, a convolution block $N$ can replace the standard convolution kernels $S$ only if two conditions are satisfied: 1) \textit{Quality Condition}: $3DRF(N) = 3DRF(S)$ for the same input tensor; 2) \textit{Compact Condition}: $3DRF(N-x) < 3DRF(S)$ if we remove a factorized kernel $x$ from $N$. The former ensures the effectiveness of $N$ with regards to its learning capacity, while the latter guarantees its optimality in terms of computation efficiency. The rule helps us unify the previous construction of the convolution block, as well as inspires us to build a new convolution blocks and one efficient factorized kernel.

\vspace{2pt}
\textbf{Unifying Existing Convolution Blocks} 
This section shows that the previous four convolution blocks follow the \textit{Rule of Kernel Replacement}: they have the same \textit{3DRF} as the standard convolutions and they are already in the compact form that cannot be further simplified.
Figure~\ref{fig:my_label} depicts the \textit{3DRF} for a standard convolution block ($S$) and four previously explored convolution blocks ($A$-$D$), in their spatial and channel dimensions.  
As shown in Figure~\ref{fig:my_label} ($S$), the \textit{3DRF} spatial size \textit{$3DRF_1^w$} for $S$ is 3 for one standard convolution and $3DRF_2^w$ is 5 when two standard convolutions are packed together in the block.
The \textit{3DRF} channel dimension $3DRF_k^c$ for $S$ equals the number of the input channels to the block. 

Convolution block $A$ (adopted by Xception~\cite{xception}) and $B$ (applied in MobileNet~\cite{mobilenet_2017_howard}) follow a similar structure. Both $A$ and $B$ successfully maintain the same \textit{3DRF} with that of $S$ with one standard kernel. 
Specifically, the spatial coverage is managed by DW
\footnote{Definitions of factorized kernels like DW can be found in the Related Work Section.}
and channel coverage is taken care of by PW, which communicates the information among all input channels. 
Convolution block $C$ (used in clcNet~\cite{zhang2018clcnet}) and $D$ (utilized by ShuffleNet~\cite{shufflenet_zhang_2017}), on the other hand, achieve the same \textit{3DRF} with that of $S$ with two standard kernels. 
Take block $C$ (shown in Figure~\ref{fig:my_label} ($C$)) as an example, one combination of GC, Interlace, and GPW, can perceive the same spatial region but only half of the entire input channels, compared to a standard convolution kernel. 
But with one extra GC+Interlace+GPW, the channel dimension gets full coverage. 
Thus, the \textit{3DRF} is the same for the block with  (GC+Interlace+GPW) * 2 and two standard convolutions. 
The proof of the compactness for four convolution blocks is omitted, but it is clear from the plot that if we remove any of the factorized kernels, the \textit{3DRF} cannot be maintained.

\begin{figure}[t]
\centering
    \includegraphics[width=\columnwidth]{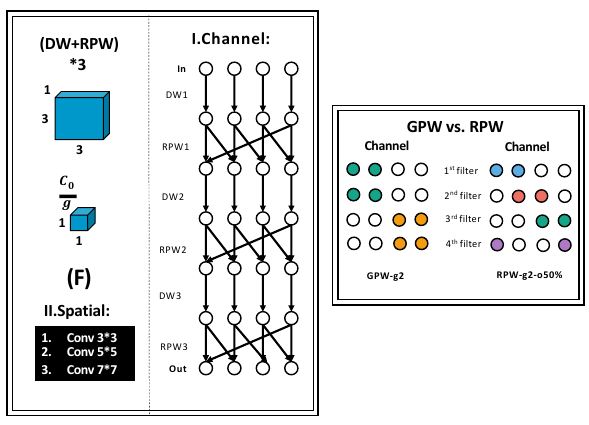}
    \vspace{-10pt}
    \caption{\textbf{Left}: DW+RPW convolution block design. \textbf{Right}: Comparison of RPW kernel with GPW kernel. Note that in RPW, adjacent filters overlap in channel dimensions.}
    \vspace{11pt}
    \label{fig:new_design}
\end{figure}

\vspace{2pt}
\textbf{New Kernel Design}
Inspired by the \textit{Rule of Kernel Replacement}, we discover an unexplored convolution block and a new type of factorized kernel, shown in Figure~\ref{fig:new_design}. 
The first block includes a DW, a channel shuffle, and a GWC. The \textit{key insight} of the design is choosing a DW to capture information in the spatial dimension and using a GPW with a shuffle operation to observe full channel information. Since the PW contributes to the majority of the computations in the previous factorized design (more than 95\% FLOPs in MobileNet~\cite{mobilenet_2017_howard}), the usage of GPW to replace PW can largely reduce the computation cost, compared to blocks like (A) and (B). 

\begin{figure}[t] \small
\begin{lstlisting}[caption=Compositing RPW via PyTorch Operators., label=Compositing RPW via PyTorch Operators.]
width = int(input_channel/num_groups)
start, end, start_v, end_v= 0, width, 0, width
item_set, slice_li = set(), []
# input channel range for each kernel filter.
for fid in range(output_channel):
    item_set.add((start,end)); slice_li.append((start,end))
    start_v = end_v - int(overlap * width) 
    end_v = start_v + width
    start, end= start_v%input_channel, end_v%input_channel
# define a groupwise convolution. 
conv2D = nn.Conv2d(width*len(item_set), len(item_set), 
                   kernel_size=1, groups=len(item_set))
# forward computation.
def forward(input):
    comb_unit = []
    for idx in range(len(item_set)):
        item = slice_li[idx]
        start, end = item[0], item[1]
        if start > end and start < input_channel:
            tmp = input[:, start:, :, :]
            tmp_1 = input[:, :end, :, :]
            new_tmp = torch.cat([tmp, tmp_1], dim=1)
            comb_unit.append(new_tmp)
        else:
            comb_unit.append(input[:, start:end, :, :])
    comb_tensor = torch.cat(combined_unit, dim=1)
    return conv2D(comb_tensor)
\end{lstlisting}
\end{figure}

The convolution block we come up with composes of a DW and a Rolling-Pointwise Convolution (RPW), as shown in the left side of Figure~\ref{fig:new_design} (model $F$). The comparison between RPW and GPW is presented in the right side of Figure~\ref{fig:new_design}. Different from GPW, RPW is the new factorized convolution kernel we invented, where adjacent convolution filters partially overlap in the channel dimension. The overlapped part serves as a bridge to communicate the different channel information and allows the later kernel to observe different channels without channel shuffle. Specifically, there are two parameters that come with RPW: group number $g$ and overlap ratio $o$.
For instance, RPW-gX-oY\% denotes each filter in the convolution kernel takes $\frac{1}{X}$ number of input channels, while adjacent filters in RPW have y\% overlap in their consumed channels.
The newly designed block outperforms previous designs in accuracy, memory and computation efficiency, which are detailed in our evaluation.


    


\textbf{Implementation of New Kernel Design} 
To implement the new rolling-pointwise convolution, we introduce two kind of implementation by compositing the existing Pytorch Operators.
First, we can first extract the corresponding channels and concatenate them together. We will leverage the existing Pytorch operators, such as tensor slicing, concatenation, and standard group convolution. There are several steps, as shown in Listing~\ref{Compositing RPW via PyTorch Operators.}. 
The second type of design is to let the convolution iterate through the input channel. 
The second implementation circumvents the ``huge'' concatenated tensor in the above implementation by applying convolution operation before concatenating. 
One major key insight is that the computation on the large concatenated tensor can be decomposed into the more effective computation on a set of small tensors. Instead of simply combining all the extracted features maps, we can pre-build a set of lightweight convolutions, each of which will generate the feature map for only one kernel filter. Finally, we concatenate these output feature map together. While this solution can largely overcome the third problem of the above channel-stack implementation, it is still hindered by the excessive inefficient Pytorch operations and lack of parallelization.

\vspace{-6pt}
\section{Evaluation} \label{sec:eval}
To validate the effectiveness of the architecture optimizer, we run comprehensive experiments on the state-of-the-art CNN models (VGG16 and VGG19~\cite{vgg}, MobileNet~\cite{mobilenet_2017_howard} and ResNet50~\cite{he2016deep}. The major reason of choosing these CNN models are 1) VGG16 and VGG19 are two most classic CNNs with linearly stacked layers; 2) MobileNet is the representative lightweight model with DW+PW convolution block; 3) ResNet50 is the representative model with the non-linearly stacked layers (residual connections).

\textbf{Dataset:} We use CIFAR-10 (CIFAR-100)~\cite{cifar10} and ImageNet~\cite{deng2009imagenet} dataset for evaluation. CIFAR-10 consists of 60,000 32$\times$32 colour images in 10 classes, with 6,000 images per class. CIFAR-100 dataset is just like the CIFAR-10, except it has 100 classes containing 600 images each. ImageNet is a large dataset of over 14 million images with up to 1,000 output classes, and it is mainly used for computer vision research, such as image classification.

\textbf{Training Settings:} We follow the conventional settings~\cite{li2016pruning} for training and testing on CIFAR-10 and CIFAR-100: learning rate starts from 0.1 and decays by the factor of 0.1 after 150 and 250 epochs, with 350 epochs in total. 
We adopt SGD with 0.9 momentum and 5e-4 for the weight decay. 
We apply normalization for the input image with (0.491, 0.482, 0.446) for each RGB channel as the mean and (0.247, 0.243, 0.261) for standard deviation, respectively. 
And we select two state-of-the-art Pytorch CNNs implementations on CIFAR-10
and CIFAR-100
, respectively.
For ImageNet, we use the official Pytorch implementations
~\footnote{\url{github.com/pytorch/examples/tree/master/imagenet}} 
and choose learning rate starts with 0.1 with total 120 epochs. We adopt SGD with 0.9 momentum and 1e-4 weight decay. 
We also apply normalization for the input image with (0.485, 0.456, 0.406) for each RGB channel as the mean and (0.229, 0.224, 0.225) for standard deviation. 
We select the pre-trained model as the baseline from Pytorch official website.

\subsection{Stage-Level Organizer} 
\label{sec:organizer}
This experiment aims to demonstrate the effectiveness of our stage-level organizer. Specifically, we first use CIFAR-10 and CIFAR-100 for detailed analysis, and further leverage ImageNet to show our design applicability and scalability towards the challenging state-of-the-art large dataset.
Table~\ref{tab:Performance comparison (CIFAR-10) between original CNNs and reorganized structures.} 
exhibits the performance of various CNNs optimized by the stage-level organizer, including computation complexity (MFLOPs), parameter size, and accuracy.
It is clear that the stage-level organizer can improve the accuracy of various state-of-the-art CNN models. 
On CIFAR-10 and CIFAR-100, stage-level organizer improves the accuracy of four evaluated models by 1.18\% and 1.90\% on average, while reducing model parameters by 54.15\% and 32.33\% on average, respectively.
We also notice on the more complicated model, such as ResNet50, the accuracy improvement is notable (2.04\% on CIFAR-10 and 0.86\% on CIFAR-100). 
The original ResNet50 model has 4 stages. Each stage contains \{3, 4, 6, 3\} bottleneck blocks respectively. Following the iterative optimization steps, the organizer moves the last two blocks from the third stage to the first stage and the last block from the last stage to the second stage to generate an optimized ResNet50 containing \{5, 5, 4, 2\} blocks in each stage. By improving the total $\Delta 3DRF$, this optimized architecture gets both higher accuracy and fewer model parameters. In addition, on the lightweight MobileNet model, which has factorized kernel designs (DW+PW) with the smallest number of parameters, our stage-level organizer also achieves a notable performance improvements (1.38\% on CIFAR-10, and 5.47\% on CIFAR-100). This is because our organizer finds five convolutions---four from the fourth stage and one from the last stage---which suffer from small $\Delta 3DRF$. By moving these convolutions to the first and second stage, we get a new architecture contains \{4, 4, 2, 2, 1\} convolutions in each stage, which offers a more efficient architecture in terms of less model parameters and higher accuracy.
On the challenging ImageNet, our stage-level organizer can still effectively reduce the number of model parameters (up to 16.7\%), meanwhile improving the testing accuracy (up to 0.58\%) compared with baseline models. 
\begin{table}[t]
    \caption{Performance comparison (CIFAR-10) between original CNNs and reorganized structures.}
    \vspace{-5pt}
    \label{tab:Performance comparison (CIFAR-10) between original CNNs and reorganized structures.}
    \centering
    \scalebox{1}{
    \begin{tabular}{ccrcc}
        \toprule
         Network & MFLOPs & Param. & Acc. (\%) & $\Delta 3DRF$\\
        \midrule 
        \midrule
        VGG16 & 310 & 14.73M & 92.64 & -\\
        \textbf{VGG16-opt} & 370 & \textbf{5.10M} & \textbf{92.95} & 2.30\\
        \midrule
        VGG19 & 400 & 20.04M & 91.91 & -\\
        \textbf{VGG19-opt} & 490 & \textbf{8.09M} & \textbf{92.89} & 3.13\\
        \midrule
        MobileNet & 50 & 3.22M & 90.67 & -\\
        \textbf{MobileNet-opt} & 50 & \textbf{1.13M} & \textbf{92.05} & 3.94 \\
        \midrule
        ResNet50 & 1,300 & 23.52M & 93.75 & -\\
        \textbf{ResNet50-opt} & 1,310 & \textbf{17.24M} & \textbf{95.79} & 0.76\\
        \bottomrule
    \end{tabular}}
    \vspace{-10pt}
\end{table}

\begin{table}[t]
    \caption{Performance comparison (\textbf{CIFAR-100}) between original CNNs and reorganized structures.}
    \label{tab:Performance comparison (CIFAR-100) between original CNNs and reorganized structures.}
    \centering
    \vspace{-5pt}
    \scalebox{1}{
    \begin{tabular}{ccrcc}
        \toprule
        Network & MFLOPs & Param. & Acc. (\%) & $\Delta 3DRF$\\
        \midrule
        \midrule
        VGG16 &  330 & 34.02M & 72.93 & - \\
        \textbf{VGG16-opt} & 390 & \textbf{24.39M} & \textbf{74.64} & 2.30\\
        \midrule
        VGG19 & 420 & 39.33M & 72.23 & -\\
        \textbf{VGG19-opt} & 500 & \textbf{27.38M} & \textbf{74.00} & 3.13\\
        \midrule
        MobileNet & 50 & 3.32M & 65.98 & -\\
        \textbf{MobileNet-opt} & 50 & \textbf{1.23M} & \textbf{71.45} & 3.94 \\
        \midrule
        ResNet50 & 1,310 & 23.71M & 77.39 & -\\
        \textbf{ResNet50-opt} & 1,380 & \textbf{21.89M} & \textbf{78.25} & 0.76\\
        \bottomrule
    \end{tabular}}
    \vspace{-10pt}
\end{table}

\begin{table}[t]
    \caption{Performance comparison (\textbf{ImageNet}) between original CNNs and reorganized structures.}
    \label{tab: Performance comparison (ImageNet) between original CNNs and reorganized structures}
    \centering
    \scalebox{1}{
    \begin{tabular}{ccrcc}
        \toprule
        Network & MFLOPs & Param. & Acc. (\%) & $\Delta 3DRF$\\
        \midrule
        \midrule
        VGG16 & 15,500 & 138.36M & 71.59& -\\\
        \textbf{VGG16-opt} & 16,900 & \textbf{133.82M} & \textbf{72.17} & 0.39\\
        \midrule
        VGG19 & 19,670 & 143.67M & 72.38& -\\\
        \textbf{VGG19-opt} & 21,060 & \textbf{141.34M} & \textbf{72.61} & 1.09\\
        \midrule
        MobileNet &  580 & 4.23M & 70.60& -\\\
        \textbf{MobileNet-opt} & 570 & \textbf{3.52M} & \textbf{71.05} & 2.59\\
        \midrule
        ResNet50 & 4,120 & 25.56M & 76.15& -\\
        \textbf{ResNet50-opt} & 4,130 & \textbf{23.67M} & \textbf{76.56} & 0.47\\
        \bottomrule
    \end{tabular}}
\end{table}

\subsection{Kernel-Level Decomposer}
This experiment aims to demonstrate the benefits of our brand-new kernel design. We first use VGG16-opt (with stage-level optimization) on CIFAR-10 for a detailed study.
We further highlight our new kernel scalability by applying it towards the complicated ResNet50-opt model on ImageNet. Table~\ref{tab:factorized_kernel VGG16} shows that our new convolution block based on rolling-channel design achieve a better balance between the model efficiency and the prediction accuracy on VGG16-opt on CIFAR10, in contrast to DW+PW factorized kernel design. We tried three different group numbers $g$ (2, 4, 8), as well as two overlapping ratios $o$ (33\%, 50\%). 
Our model with DW+RPW-g2-o50\% achieves a better accuracy compared to the high-performance DW+PW model while saving about 40.0\% FLOPs and 40.5\% parameters. 
With an increase in the group number, we observe a significant reduction in both computational cost and parameter usage, along with a slight degradation in prediction accuracy. 
This aligns well with our expectation that the group number $g$ determines the number of input channels that GPW/RPW would take, and thus also decides the number of computations and parameters of the model.

We also notice that our new convolution block design consistently outperforms with the ones without overlap ($o$) under the same number of groups ($g$). For example, our new design (DW+RPW-g4-o33\%) outperform DW+RPW-g4 with 3.56\% better accuracy. Under the settings with same number of group in RPW, such as DW+RPW-g2-o33\% vs. DW+RPW-g2-o50\%, the latter with higher overlap ratio offers higher accuracy, indicating the effectiveness of overlapping channels to improve model accuracy.
\begin{table}[t]
    \caption{Kernel-level design (\textbf{CIFAR-10}) on VGG16-opt.}
    \label{tab:factorized_kernel VGG16}
    \centering
    \scalebox{1}{
    \begin{tabular}{cccc}
        \toprule
        Network & MFLOPs & Param. & Acc.(\%)\\
        \midrule
        \midrule
        Baseline & 370 & 9.64M & 92.95\\
        \midrule
        DW+PW & 50 & 1.11M & 92.12\\
        \midrule
        DW+GPW-g2 & 30 & 0.67M & 92.35\\  
        \midrule
        DW+GPW-g4 & 20 & 0.36M & 88.05\\
        \midrule
        DW+GPW-g8 & 10 & 0.20M & 86.41\\
        \midrule
        \textbf{DW+RPW-g2-o33\%} & \textbf{30} & \textbf{0.66M} & \textbf{92.52}\\
        \midrule
        \textbf{DW+RPW-g2-o50\%}& \textbf{30} & \textbf{0.66M} &
        \textbf{92.70}\\  
        \midrule
        \textbf{DW+RPW-g4-o33\%} & \textbf{20} & \textbf{0.36M} & \textbf{91.61}\\
        \midrule
        \textbf{DW+RPW-g4-o50\%} & \textbf{20} & \textbf{0.36M} &
        \textbf{91.59}\\
        \midrule
        \textbf{DW+RPW-g8-o33\%} & \textbf{10} & \textbf{0.20M} & \textbf{89.86}\\ 
        \midrule
        \textbf{DW+RPW-g8-o50\%} & \textbf{10} & \textbf{0.20M} & \textbf{90.19}\\   
        \bottomrule
    \end{tabular}}
\end{table}
\section{Conclusion}
\label{conclusion}
In this paper, we propose \textit{3D-Receptive Field (3DRF)}, an interpretable and easy-to-compute metric to guide the search of CNN designs. 
To illustrate the usefulness of \textit{3DRF}, We build an optimizer and improve the CNN structure at the stage and kernel level. The stage-level optimization target at reducing the model structural redundancy by improving the kernel organization, while the kernel-level optimization improve the individual kernel design by reducing the number of parameters without much compromising the model accuracy. 
Experiments show models generated by our optimizer achieve higher efficiency and accuracy compared with state-of-the-art CNNs. 

\section{Acknowledgment}
This work was supported in part by NSF 1925717. 
Use was made of computational facilities purchased with funds from the National Science Foundation (OAC-1925717) and administered by the Center for Scientific Computing (CSC). The CSC is supported by the California NanoSystems Institute and the Materials Research Science and Engineering Center (MRSEC; NSF DMR 1720256) at the University of California, Santa Barbara.

\bibliographystyle{ACM-Reference-Format}
\balance
\bibliography{reference}

\end{document}